\documentclass[twocolumns,12pt]{article}
\usepackage{arxiv}

\pdfoutput=1 
\usepackage[utf8]{inputenc}
\usepackage{bm}
\usepackage{graphicx}
\usepackage{hyperref}
\usepackage{amsfonts}
\usepackage{amssymb,amsmath}

\title{Exploiting video sequences for unsupervised disentangling in generative adversarial networks}

\author{
  Facundo Tuesca\thanks{This preprint is the result of the work done for
  the undergraduate dissertation of F. Tuesca supervised by L.C. Uzal and presented in June 2018.}\\
  Facultad de Ciencias Exactas, \\
  Ingeniería y Agrimensura, UNR \\
  Argentina.\\
   \And
 Lucas C.~Uzal \\
  CIFASIS, French Argentine International \\
  Center for Information and Systems Sciences, \\
  UNR-CONICET, Argentina. \\
  \texttt{uzal@cifasis-conicet.gov.ar} \\
 }

\begin{document}
\maketitle

\begin{abstract}
In this work we present an adversarial training algorithm that exploits correlations in video to learn --without supervision-- an image generator model with a disentangled latent space. 
The proposed methodology requires only a few modifications to the standard algorithm of Generative Adversarial Networks (GAN) and involves training with sets of frames taken from short videos. We train our model over two datasets of face-centered videos which present different people speaking or moving the head: VidTIMIT and YouTube Faces datasets.
We found that our proposal allows us to split the generator latent space into two subspaces.
One of them controls \textit{content} attributes, those that do not change along short video sequences. For the considered datasets, this is the identity of the generated face. The other subspace controls \textit{motion} attributes, those attributes that are observed to change along short videos. We observed that these motion attributes are face expressions, head orientation, lips and eyes movement. The presented experiments provide quantitative and qualitative evidence supporting that the proposed methodology induces a disentangling of this two kinds of attributes in the latent space.
\end{abstract}


\section{Introduction}
Generative Adversarial Networks (GANs) \cite{gan2014} are models based on neural networks which have been shown to be very effective at artificial image generation \cite{Reed2016, Karras2017, Zhang2018}. 
An interesting property of GANs is that the generator, once trained, is a deterministic map between a vector $z$ and the generated sample (an image in our case). All the data required for this transformation is contained in the input $z$ vector and in the fixed parameters (learned weights and biases) of the generator: there are no stochastic processes at work. In this context, it is of interest to study the latent space learned by the generator and how the components of $z$ control the semantic content of the generated image. In the seminal work of Radford et al. \cite{Radford2015} there is a first insight into the semantic relationship between pairs of images generated by vectors $z$ close to each other and the image transformations caused by moving $z$ in different directions.

In this paper we propose an adversarial learning setup that allows us to condition the latent space that will be learned. We divided the latent space into subspaces that capture a semantic decoupling of high level features of the generated images. The proposed methodology does not require any labeled data and it is based only on temporal correlations inferred from video datasets. In particular, we show that using a video dataset of people talking we can subdivide the latent space into two subspaces: one for \textit{content} attributes (e.g. identity of the face), and the other for \textit{motion} attributes (e.g. face expression). We also show different properties of the induced latent space, and how it is affected by the choice of the considered datasets.

\section{Methodology}
\begin{figure*}[h]
 \centering
 \includegraphics[width=\textwidth,keepaspectratio=true]{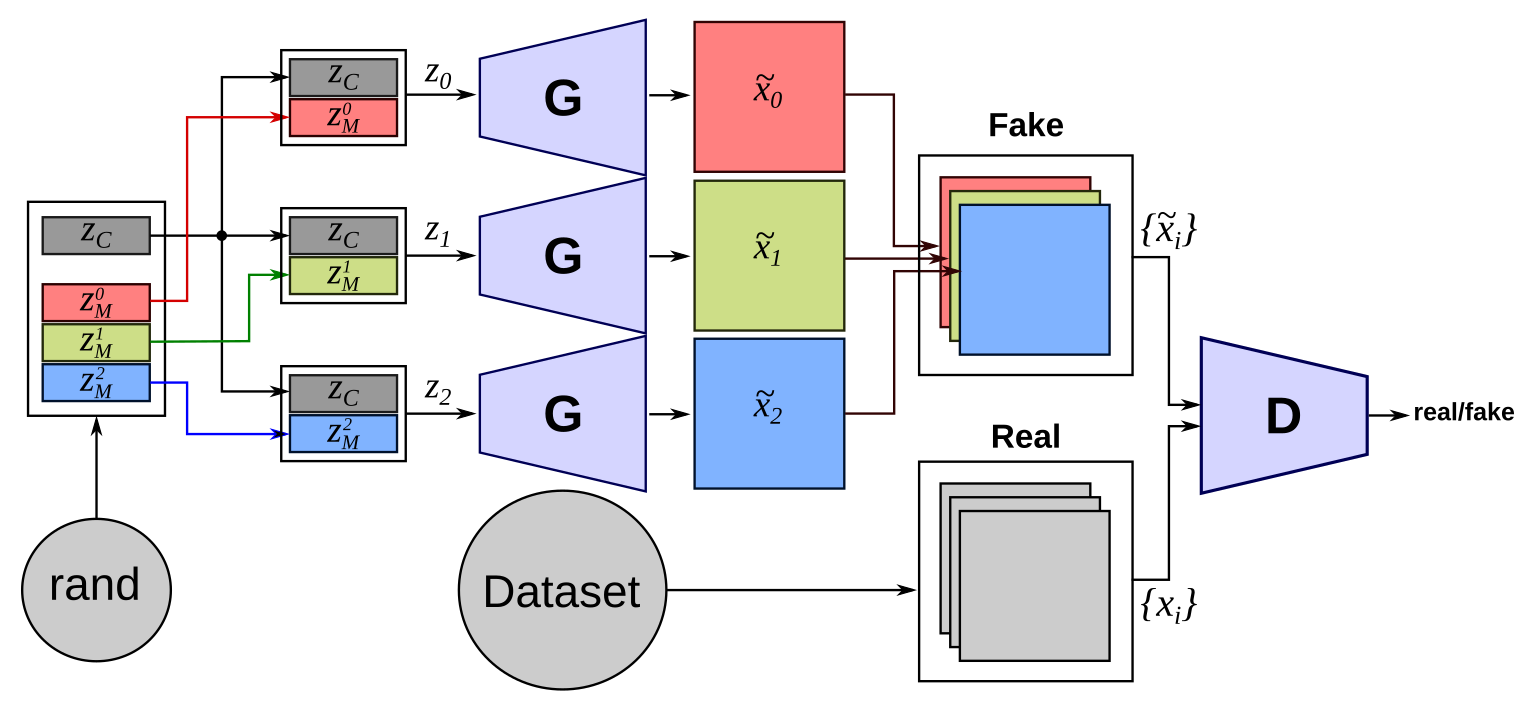}
\caption{Diagram of connections for the proposed methodology. Discriminator $D$ takes groups of $3$ frames (generated or real) stacked along the channel axis. Generator $G$ takes a random vector $z$ and outputs a single image as usual. We composed each $z$ with two parts: $z_C$ is the fixed subvector that controls the time-invariant attributes of the images, while the set $\{z^0_M,z^1_M,z^2_M\}$ contains the subvectors associated to attributes that may change with time. $\{\tilde{x}_i\}$ is the set of generated images, and $\{x_i\}$ is the set of frames of the same short video from the dataset.}
 \label{fig:model1}
\end{figure*}

The proposed methodology (Fig. \ref{fig:model1}) can be implemented with only a few modifications to the standard adversarial training algorithm for generating images. Generator $G$ remains the same and therefore samples a single image for each single input vector $z$. On the other hand, discriminator $D$ should be modified in order to process a set of $n$ images at the same time. This set could contain real or fake images. If the set contains real images, they are taken randomly from the frames of one single short video clip of the dataset. On the contrary, if the set contains fake images, they come from applying $G$ to a set of $z$ vectors designed as follows: We subdivide the $z$ vector in two parts: $z = (z_C, z_M)$, where $z_C$ remains fixed for the $n$ elements in the set while $z_M$ varies. In other words, a random $z_C$ is sampled once per set while $z_M$ is resampled for each element of the set. $z_C$ is intended to hold the static (or content) information of a frame (i.e. in a video of a person talking, this vector is expected to hold the information that determines the identity of the person). On the other hand, $z_M$ will hold the information related to dynamic (or motion) attributes: the high level features of a frame that can change with time (such as the expression of the face).

Our proposal, can be stated in terms of a redefinition of the value function $V(G,D)$ for the adversarial training as this minmax game:
\begin{equation}\label{eq:minmax}
\begin{split}
\min_G \max_D V(D,G) =\ & \mathbb{E}_{\{x_i\}\sim p_{data}(\{x_i\})}[\log D(\{x_i\})]+\\
& \mathbb{E}_{\{z_i\}\sim p^*_z(\{z_i\})}[\log (1 - D(\{G(z_i)\}))]
\end{split}
\end{equation}
where $\{x_i\}$ is a set of $n$ images ($i=1\ldots n$) sampled randomly from a single video sampled in turn from a dataset of short videos. $\{z_i\}$ is the set of $n$ random vectors with components sampled from uniform distribution in $[-1,1]$ with a part $z_C$ fixed for all $i$, and a part $z_M$ resampled for each $i$, as explained previously. This sampling procedure for sets $\{z_i\}$ is captured by $p^*_z(\{z_i\})$ in Eq. \ref{eq:minmax}.  

In this work we consider $n=3$ and we simply expand the number of input channels of $D$ from $c$ channels (typically $c=3$ for RGB images) to $c\times n$ channels to cope with multiple images. Thus it was possible to reuse the available architectures for $D$ and $G$. 
Furthermore, we considered grayscale images ($c=1$), therefore $D$ has 3 input channels as usual, and $G$ has a single output channel.

The present formulation (Eq. \ref{eq:minmax}) was designed to condition $G$ to learn a latent space in which $z_C$ determines the constant elements of the image, while $z_M$ determines the elements that can change with time.

\section{Related work}
Our proposed model and all the related work are inspired by the seminal work of Radford et al. \cite{Radford2015}.
In particular, we refer to their analysis of the emerging properties of latent space after adversarial training.
They showed that some specific directions in the latent space capture disentangled high-level features of the generated data. 
For example, with a dataset of faces, they found directions that allowed them to change head rotation, smile, glasses, etc., without significantly altering face identity. 
However, finding these meaningful directions requires a carefully search over a huge sample of generated data. For this reason many papers have focused on handling or delimiting these directions by somehow conditioning GAN training.
This can be achieved by exploiting class labels with supervised datasets as the encoder-decoder architecture in \cite{Lample2017}.
Also a significant amount of disentangling can be achieved with a purely unsupervised setup like in InfoGAN \cite{Chen2016}.
All these approaches work on single images and do not exploit video sequences as in our proposal. 

There are also some attempts to exploit video datasets for disentangling motion attributes from content attributes for video generation. In this context, there are two similar approaches, MoCoGAN \cite{Tulyakov2017} and TGAN \cite{Saito2017}. In both cases, a model based on GAN generate videos using an image generator applied to a sequence of input vectors $z$. In order to generate semantically valid videos, the $z$ vector is subdivided in two: a content vector $z_C$ that determines the static characteristics of a frame, and a movement vector $z_M$ that determines how those static elements move inside the frame. This is quite similar to the model proposed here, except for the fact that the vectors $z_M$ have to be generated with a second generator: a recurrent neural network in MoCoGAN, and a second generator of fixed length vectors in TGAN. Also, in MoCoGAN there are two discriminators: one for static images and another for groups of images (videos). This means that these approaches achieve a disentangling of the latent space, but require a second generator to know which values of $z_M$ generate valid images and in which directions $z_M$ must move in order to generate a valid video. MoCoGAN also requires a second discriminator to learn to generate valid videos which is a different objective than ours.

Another (less related) approach is considered in \cite{Vondrick2016}, in which a video generator is learned by using two separate generators: one for the video background (a static image) and another one for the foreground (a series of images representing a moving object). Because these two generators use the same initial vector $z$, there is not a direct conditioning of the latent space: we only know that both generators are free to use all components in the vector in order to create the new image. This method relies on the strong assumption that videos have an static background over which some objects (foreground) evolve.

\section{Experimental}
We trained the model with two different datasets of videos of human faces: VidTIMIT \cite{vidtimit} and YouTube Faces \cite{youtubefaces}. These datasets have the following characteristics:
\begin{itemize}
\item VidTIMIT is a dataset of videos of 43 different people reciting short sentences and moving their heads. The recording was done in high quality, with a static background which remained the same for all videos.
\item YouTube Faces is a dataset of videos of $1595$ different people, downloaded from YouTube. These videos do not have anything in common, except that they contain one or multiple faces. The backgrounds are diverse and the image quality greatly varies between samples.
\end{itemize}
The differences between these two datasets allowed us to analyze the influence that the training data has over how our model conditions the latent space.

First of all, data were preprocessed by cropping and aligning faces. To this end, we designed an algorithm that exploits temporal continuity of face location and bounding box size along video frames. 
Concretely, the algorithm applies corrections to the coordinates and size of the detected bounding boxes by smoothing the corresponding time series.
To this end, we fit a spline to the time series of coordinates and sizes obtained from a standard face detection method applied to each single frame. 
With these corrections, we get a smooth movement of the bounding box along frames.
Also, this procedure allows us to recover (by interpolation) non detected bounding boxes.

\begin{figure}[h]
 \centering
 \includegraphics[height=2cm,keepaspectratio=true]{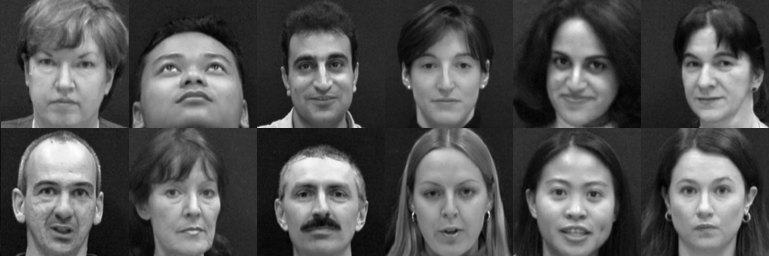}

 \caption{Random image samples of the VidTIMIT dataset after cropping.}
 \label{fig:vidtimit}
\end{figure}

\begin{figure}[h]
 \centering
 \includegraphics[height=2cm,keepaspectratio=true]{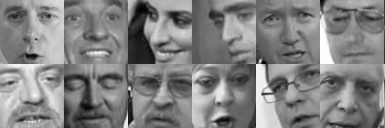}
 
\caption{Random image samples of the YouTube Faces dataset after cropping.}
 \label{fig:ytfaces_cropped}
\end{figure}

We performed a very aggressive cropping of the data in YouTube Faces to minimize the background noise: a less strict cropping ended up affecting the stability of the training. Figures \ref{fig:vidtimit} and \ref{fig:ytfaces_cropped} show final image samples for each dataset.

Regarding model details, for networks $G$ and $D$ we consider the DCGAN architecture \cite{Radford2015}. We use the input channels of $D$ for allocating the set of images as explained before. We train both networks following the procedure in \cite{Radford2015}.
\section{Results and discussion}
\subsection{Image generation}
\begin{figure}[h]
 \centering
 \includegraphics[height=2cm,keepaspectratio=true]{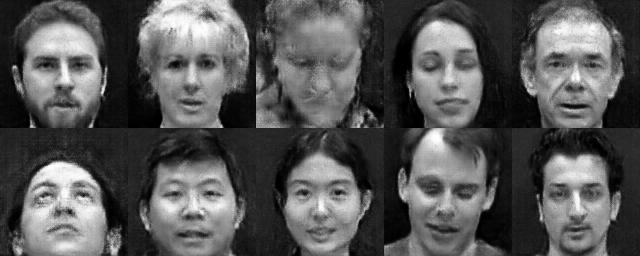}
 \caption{Faces generated by the model trained with the VidTIMIT dataset.}
 \label{fig:vidtimit_generated_sample}
\end{figure}

\begin{figure}[h]
 \centering
 \includegraphics[height=2cm,keepaspectratio=true]{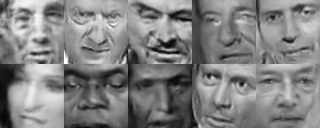}

 \caption{Faces generated by the model trained with the YouTube Faces dataset.}
 \label{fig:ytfaces_generated_sample}
\end{figure}

We first show that network $G$ trained with our method learns to generate fake images from random input $z$ as in standard GAN formulation.
In particular, Fig. \ref{fig:vidtimit_generated_sample} and Fig.
\ref{fig:ytfaces_generated_sample} show random fake images for each considered dataset.
We found that faces generated for VidTIMIT can always be traced back to one of the 43 identities present in the training data. This is an expected overfitting behaviour given the small number of identities available. For YouTube Faces dataset it is not straightforward to identify if generated faces are present in the training set and it is more likely to find generated faces as a product of high level identity interpolation as reported for celebA dataset \cite{Radford2015}.

\subsection{Semantic disentangling}
\begin{figure*}[h] 
 \centering  
\includegraphics[height=4cm,keepaspectratio=true]{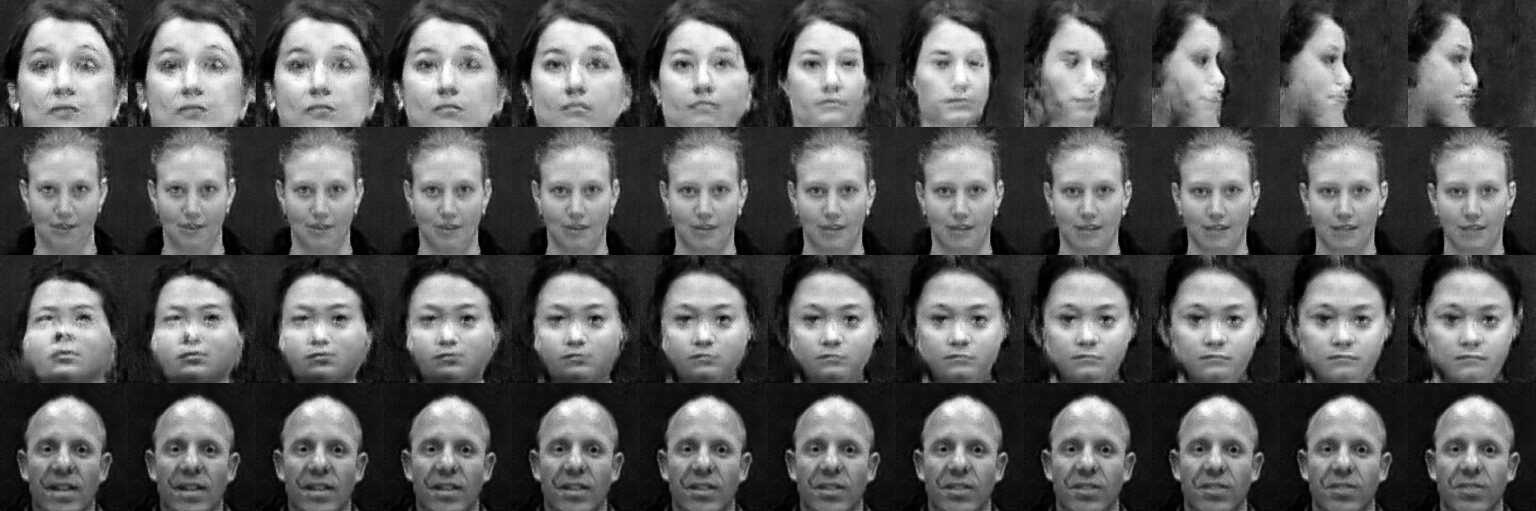}
\caption{Semantic disentangling results for VidTIMIT dataset. Each row corresponds to a different $z_C$ fixed for all columns, and the columns corresponds to $z_M$ ranging the intervals $[-z*_M, z*_M]$ for random $z*_M$ (resampled for each row). Notice that the identity of faces remains unchanged along each row while face expressions and position significantly change.}
 \label{fig:vidtimit_differentIdentity_differentDelta}
 \end{figure*}

\begin{figure*}[h] 
 \centering  
\includegraphics[height=4cm,keepaspectratio=true]{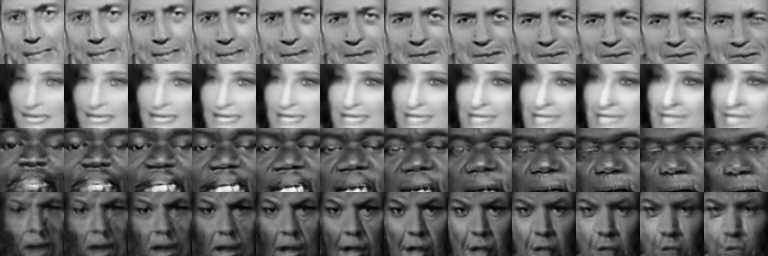}
 \caption{Semantic disentangling results for YouTube Faces dataset. We follow the same procedure as in Fig. \ref{fig:vidtimit_differentIdentity_differentDelta}. Notice that, again, the identity of faces remains unchanged along each row while face expression and position significantly change.}
 \label{fig:ytfaces_differentIdentity_differentDelta}
\end{figure*}

We now show how the model learns a latent space in which $z_C$ controls the identity of the generated image, while $z_M$ determines its expression. Figures \ref{fig:vidtimit_differentIdentity_differentDelta} and \ref{fig:ytfaces_differentIdentity_differentDelta} show the result of leaving $z_C$ fixed for each row while smoothly changing the values of $z_M$, for both datasets.

The fact that identities stay the same for faces in the same row (i.e. for fixed $z_C$) while some other factors evolve (attributes like lip or head movement), suggest that there is a decoupling of the identity of the faces from their expressions in the latent space representation.

\subsection{Inner structure of the latent space}
\begin{figure*}[h] 
 \centering  
\includegraphics[height=3cm,keepaspectratio=true]{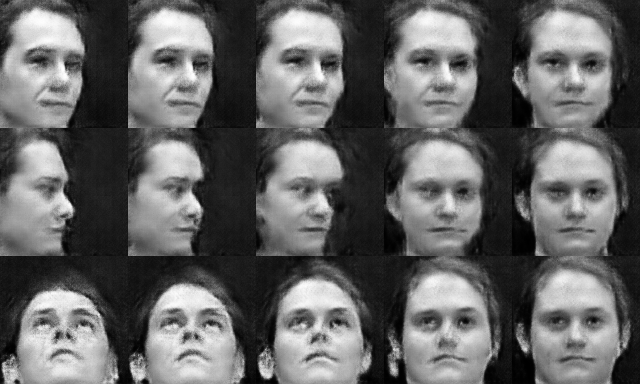}

 \caption{Latent space analysis for the model trained on VidTIMIT dataset. All rows correspond to the same $z_C$, and the columns corresponds to $z_M$ ranging the intervals of the form $[-z*_M, z*_M]$ for random $z*_M$ (resampled for each row). With this particular $z_C$, the movements generated by using different $z_M$ are only of one type (head movements).}
 \label{fig:vidtimit_sameIdentity_differentDelta}
\end{figure*}

An interesting question is how identities and motion attributes are distributed in the latent space.
We observed that, for the model trained with VidTIMIT, the expressions generated by varying $z_M$ but leaving $z_C$ fixed are always of the same type. They are either head movements or mouth/eye movements, but we did not find any single $z_C$ with both kind of movements when varying $z_M$.
This can be seen in Fig. \ref{fig:vidtimit_sameIdentity_differentDelta}, where the generated images only contain head movements.

\begin{figure*}[h]  
\centering  
 \includegraphics[width=12cm,keepaspectratio=true]{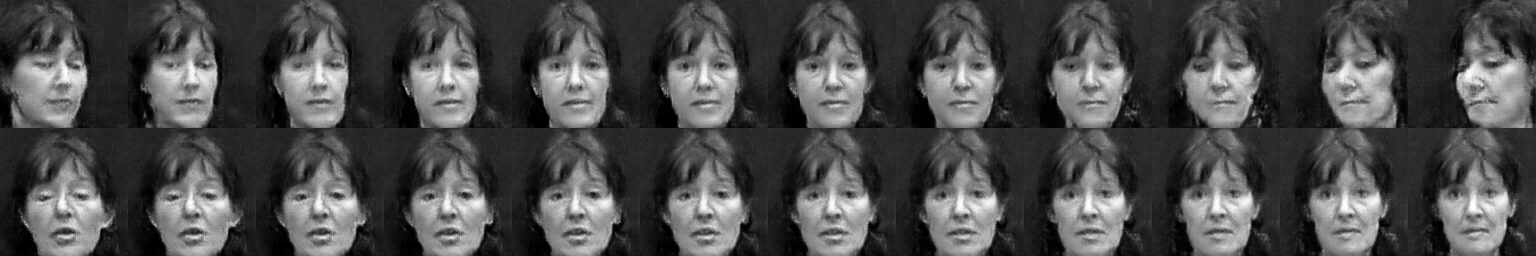}

 \caption{Latent space analysis for the model trained on VidTIMIT dataset. In this case, each row corresponds to a different $z_C$ that generates the same identity while both rows correspond to the same movement through the $z_M$ subspace. Results show different motionface attributes evolving in each row implying that --for this dataset-- the generated expression depends partially on the $z_C$ used for this dataset.}
 \label{fig:vidtimit_sameIdentity_different_zi}
\end{figure*}

Also for VidTIMIT dataset, we found pairs of different $z_C$ which generate the same identity. 
Furthermore, the effect of modifying $z_M$ produce a different kind of expression for each $z_C$ of the same identity pair. 
For example, in Fig. \ref{fig:vidtimit_sameIdentity_different_zi} each row is generated by a
different $z_C$ that produces the same identity. 
Changing the $z_M$ values results in different types of movement for each of them: in
the first row there is head movement, whereas in the second row there is mouth and eyes movement.

This dependency between the expression and the $z_C$ can be explained as follows: 
In the VidTIMIT dataset there are videos of people moving their heads or talking (but never both at the same time). 
This explains where the separation came from. 
By representing this two kinds of movement at different $z_C$ the generator prevents from generating combinations of both that can be easily detected as fake by the discriminator as these combinations do not occur in the real samples.
On the other hand it can be argued that given the fact that there is a low number of identities and videos in VidTIMIT dataset, the model might have more capacity than necessary. 
In other words, that the model is free to learn a latent space in which the same identity can be represented by several different $z_C$ given that the amount of total identities is so low that there is no need to learn an efficient representation in the latent space.
We performed experiments where we gradually reduce $z_C$ dimension or alternatively $G$ capacity (by reducing the number of maps of each layer) and we found that eventually image quality deteriorates but the dependency between the expression and the $z_C$ does not disappear. This suggest that the observed dependency is inherited from the dataset and is not related to model capacity. 

The above hypothesis implies that for a dataset like YouTube faces (which contains both kinds of movement without an imposed separation) it is more likely that the model represents each identity with a single $z_C$ and presents no dependency between $z_C$ and face expressions. 
Effectively, for the model trained with YouTube Faces we were not able to find two distinct $z_C$ that represent the same identity. Furthermore, we found that from each single $z_C$ a larger variety of attributes can be generated by varying $z_M$.

\begin{figure*}[h]  
\centering   \includegraphics[width=12cm,keepaspectratio=true]{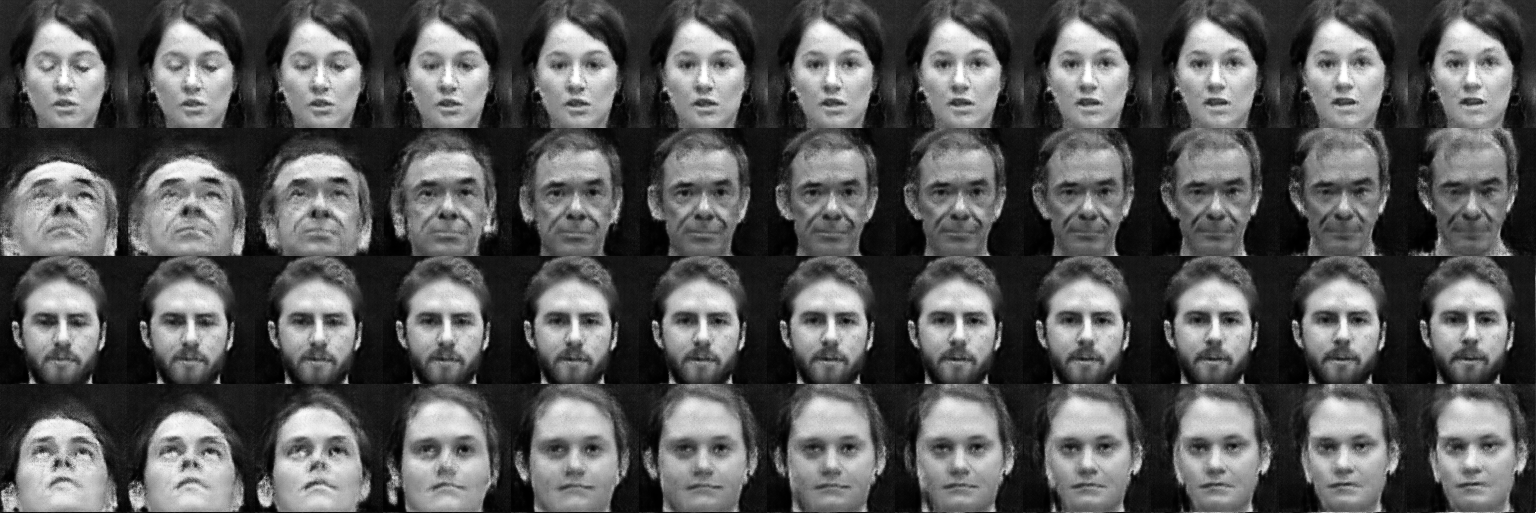}
 \caption{Disentangling failure for the model trained on VidTIMIT dataset. Each row corresponds to a different $z_C$, and the columns corresponds to $z_M$ ranging the interval $[-z*_M, z*_M]$ (using the same $z*_M$ for all rows). Notice that the same $z_M$ sequence results in different movements for each $z_C$, showing that there is not a one-to-one correspondence between $z_M$ and the generated expressions in the latent space for this dataset.}
 \label{fig:vidtimit_differentIdentity_sameDelta}
\end{figure*}

\begin{figure*}[h] 
 \centering  
\includegraphics[width=12cm,keepaspectratio=true]{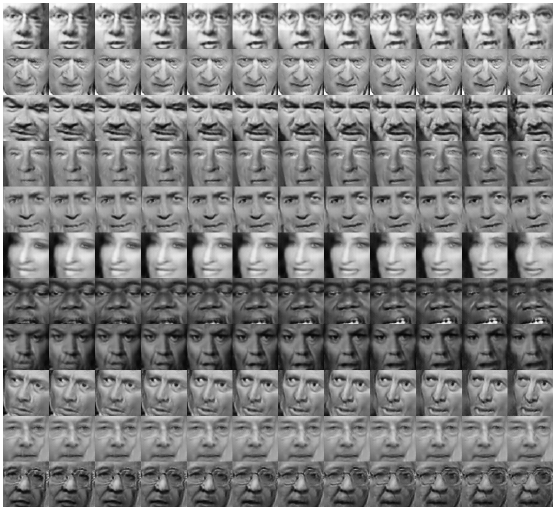}

 \caption{Disentangling success for the model trained on YouTube Faces dataset. We follow the same procedure as in Fig. \ref{fig:vidtimit_differentIdentity_sameDelta}. As opposed to the VidTIMIT case, the same $z_M$ generally produces nearly the same movement for different identities.
We invite the reader to consult the accompanying video (\url{https://youtu.be/rTSfH9ZMt68}) which shows different $z_C$ with the same $z_M$ varying with time.}
 \label{fig:ytfaces_differentIdentity_sameDelta_full}
\end{figure*}

We also studied the effect of a fixed $z_M$ sequence for different $z_C$. Ideally, we expect to find the same sequence of facial expressions observed over different face identities for a disentangled representation. 
In Fig. \ref{fig:vidtimit_differentIdentity_sameDelta} we show that with VidTIMIT, the same $z_M$ generates different expressions/movements for different identities. 
This means that the expressions determined by $z_M$ were also not consolidated in the latent space, since the same $z_M$ produces a head movement for one identity, and a mouth
movement for another.
In contrast, the model trained with YouTube Faces has learned a more disentangled representation, in which each $z_M$ generates the same expression for different identities (as shown in Fig.\ref{fig:ytfaces_differentIdentity_sameDelta_full}). 
We found that the facial expression determined by one $z_M$ is mostly the same for different $z_C$, providing evidence of a semantic decoupling between this two subspaces that expand the latent space. We invite the reader to consult the accompanying video (\url{https://youtu.be/rTSfH9ZMt68}) which shows the result of considering different $z_C$ with the same $z_M$ varying with time. This produce nearly the same sequence of expressions for different identities. This latter behavior is not explicitly conditioned by Eq. \ref{eq:minmax} but rather it is an emerging property indirectly induced by Eq. \ref{eq:minmax} and the limited capacity of the model.

\subsection{Quantitative results}
In this section we show how $z_C$ and $z_M$ affect the identity of the generated face. 
We rely on the open source face recognition library OpenFace \cite{openface} to provide quantitative results. Besides detecting faces, the library has the option to get a vector representation of the detected face with the property that the Euclidean distance in that vector space --\textit{OpenFace distance} in the following-- corresponds with a dissimilarity between the faces: a value below 1 suggest that the faces share the same identity. 

\begin{figure*}[h]
 \centering  
 \includegraphics[width=\textwidth,keepaspectratio=true]{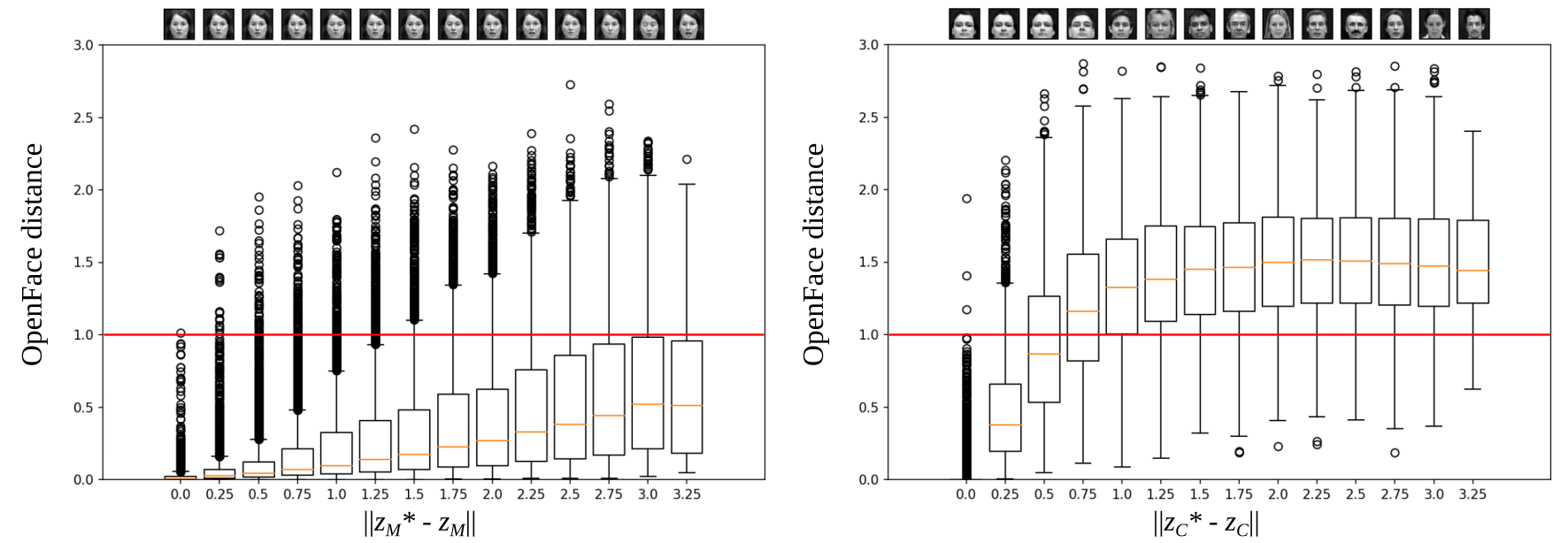}

 \caption{Identity dissimilarity based on the OpenFace distance between pairs of generated images as a function of the norm of the corresponding perturbation in the latent space (VidTIMIT dataset). A distance below 1 suggest that the faces share the same identity. This shows how moving a $z$ vector in the $z_M$ subspace generally preserves the identity for all perturbation sizes, whereas a perturbation in the $z_C$ subspace sharply changes the identity of the generated face.}
 \label{fig:vidtimit_boxplot}
\end{figure*}

\begin{figure*}[h]
  \centering  
 \includegraphics[width=\textwidth,keepaspectratio=true]{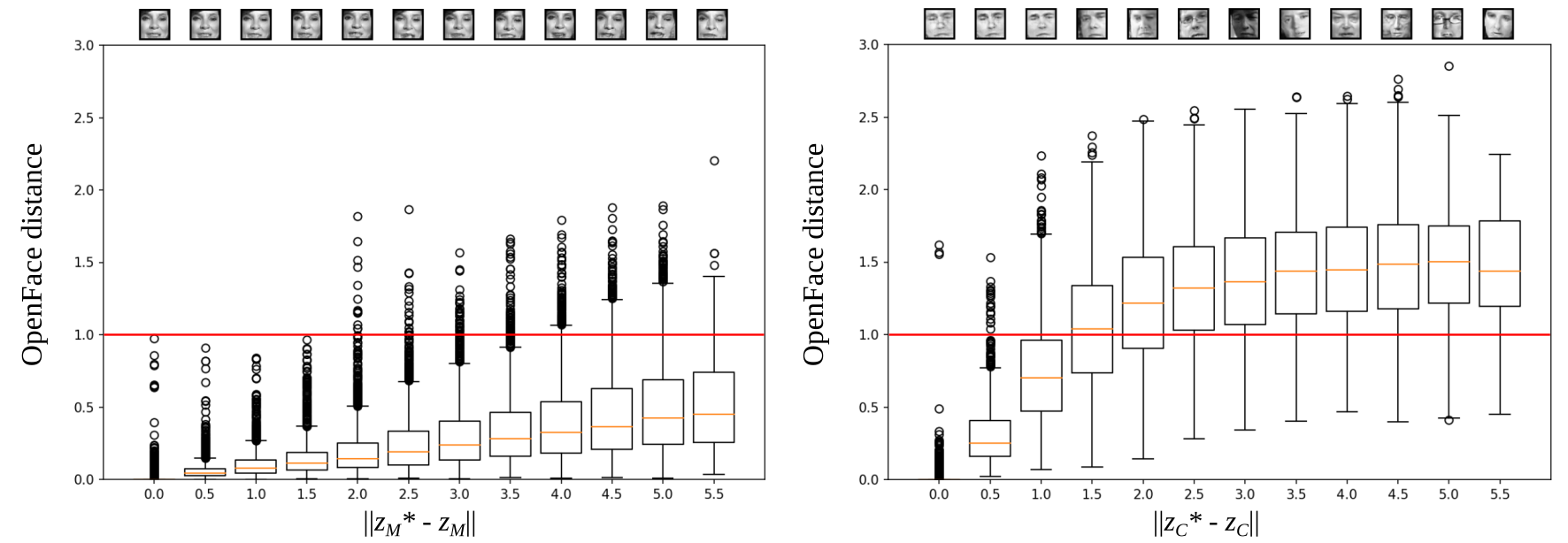}

 \caption{Same as Fig. \ref{fig:vidtimit_boxplot}, now for YouTube Faces dataset. The same behavior is observed.}
 \label{fig:ytfaces_boxplot}
\end{figure*}

In particular, we show how distances in $z_C$ and $z_M$ subspaces correlate with the OpenFace distance. 
Figures \ref{fig:vidtimit_boxplot} and \ref{fig:ytfaces_boxplot} show, for VidTIMIT and YouTube Faces respectively, a boxplot for OpenFace distances distribution versus different perturbation norms in the $z_M$ subspace (left panels) and in the $z_C$ subspace (right panels). 
In all cases, the $X$ axes expand the full range of distances observed with probability $p>0.01$ between random pairs.
It can be observed that, for both datasets, when perturbations are in the $z_M$ subspace, the identity of the generated faces does not change in most cases as the boxes\footnote{Boxes extend from the lower to upper quartile values of the data.} are clearly below the threshold of 1.
On the other hand, when perturbation is along $z_C$ subspace, the identity distance grows rapidly above the threshold. 
The generated images shown above plots correspond to a single selected example that illustrates how a face changes when its corresponding $z$ is accordingly perturbed. 
It can be seen in the right panels of both figures how the generated faces visit different identities when the OpenFace distance saturates above the threshold.


\section{Conclusions and future work}
We proposed a model based on minimal modifications to the standard GAN formulation that
can exploit correlations in video to learn without supervision a disentangled latent space for image generation.

We test our model over two datasets of face-centered videos which present different people speaking or moving the head: VidTIMIT and YouTube Faces datasets.

We found that our proposal allows us to split the generator latent space into two subspaces.
One of them controls the \textit{content} attributes, those that do not change along short video sequences. For the considered datasets, this subspace controls the identity of the generated face. The other subspace controls motion attributes, those attributes that are observed to change along short videos. For the considered datasets, these motion attributes are present in the face: expressions, head orientation, lips and eyes, etc. 

Experiments over YouTube Faces dataset showed a complete disentangling of the above mentioned attributes. Identity and facial attributes can be independently controlled. This means that we can reach generated faces with the same expressions and different identities and, alternatively, the same identity with different expressions, by moving the input vector over each latent subspace correspondingly.

We also provide a quantitative analysis over a large sample of generated images: we test the incidence on face identity caused by perturbations over each disentangled subspace. To this end we rely on a face recognition library --OpenFace \cite{openface}-- which provides a vector representation for face images that allows us to measure the similarity of identities between pair of images. This quantitative analysis provides further evidence that supports the hypothesis of a disentangling in the latent space.

Our analysis used datasets of videos of people talking, but the proposed method can be applied to any dataset in which there is at least one feature which remains fixed while the other changes. 
The disentangling of attributes is a desirable property that gives us improved control over generated images. 

Our future work is oriented to generate fake images of faces for a desired identity for which a very few real samples are available. We believe that the disentangling reached in the present work will facilitate to transfer facial expressions learned from a large dataset to a new unseen identity. 
\section{Acknowledgements}
The authors acknowledge grant support from ANPCyT PICT-2016-4374 and ASaCTeI IO-2017-00218.

\bibliography{bibliografia}
\bibliographystyle{unsrt}

\end{document}